\documentclass{article}
\usepackage{spconf,amsmath,graphicx}

\usepackage{booktabs}
\usepackage{endnotes}
\usepackage{multicol}
\usepackage{multirow} 
\usepackage{float}

\title{Fast-HuBERT: An Efficient Training Framework for Self-Supervised Speech Representation Learning}
\name{Guanrou Yang, Ziyang Ma, Zhisheng Zheng, Yakun Song, Zhikang Niu, Xie Chen$^{\ast}$ \thanks{*Corresponding author.}}
\address{MoE Key Lab of Artificial Intelligence, AI Institute, X-LANCE Lab \\
Department of Computer Science and Engineering, Shanghai Jiao Tong University\\
\texttt{\{yangguanrou, chenxie95\}@sjtu.edu.cn}}

\begin{document}
%
\maketitle

\begin{abstract}
Recent years have witnessed significant advancements in self-supervised learning (SSL) methods for speech-processing tasks.
Various speech-based SSL models have been developed and present promising performance on a range of downstream tasks including speech recognition. 
However, existing speech-based SSL models face a common dilemma in terms of computational cost, which might hinder their potential application and in-depth academic research.
To address this issue, we first analyze the computational cost of different modules during HuBERT pre-training and then introduce a stack of efficiency optimizations, which is named Fast-HuBERT in this paper. 
The proposed Fast-HuBERT can be trained in 1.1 days with 8 V100 GPUs on the Librispeech 960h benchmark, without performance degradation, resulting in a 5.2x speedup, compared to the original implementation.
Moreover, we explore two well-studied techniques in the Fast-HuBERT and demonstrate consistent improvements as reported in previous work. \footnote{The code for Fast-HuBERT training is available at https://github.com/yanghaha0908/FastHuBERT}

\end{abstract}

\begin{keywords}
Speech recognition, Self-supervised learning, Efficiency optimization
\end{keywords}

\section{Introduction}
\label{sec:intro}

SSL becomes increasingly popular as it does not require manually labeled data and can learn universal representations from a large amount of unlabeled data. 
These representations are proven to work well on a variety of downstream tasks including speech recognition \cite{e2ereview}. 
The training procedure of SSL models is typically divided into two stages. 
In the first stage, a representation model is pre-trained by learning to extract useful structural information from a large amount of unlabeled data. 
In the second stage, the representations of a frozen or tuned pre-trained model are utilized in a supervised downstream task. 
Due to their excellent generalization performance, SSL models gain widespread attention in both academia and industry and also become a popular research topic in the field of speech processing. 
There are prominent self-supervised speech representation models~\cite{baevski2020wav2vec, hsu2021hubert, chen2022wavlm, ma2022mt4ssl, meng2022cobert}, among which HuBERT~\cite{hsu2021hubert} exhibits robust performance across multiple tasks~\cite{yang2021superb} such as speech recognition~\cite{wang2023hubert}, speaker verification~\cite{chen2022large}, and emotion recognition~\cite{ma2023leveraging}. 

However, training SSL models usually requires huge computational resources for using large-scale unlabeled data and large parameter models to adequately absorb the knowledge embedded in the data. 
For example, the base model of wav2vec 2.0 requires 1.6 days of training on 64 V100 GPUs, and its large model requires 2.3 days of training on 128 V100 GPUs~\cite{baevski2020wav2vec}. 
The training of HuBERT takes even longer for its iterative re-clustering and re-training. 
The base model of wav2vec 2.0 and HuBERT consists of 95 million parameters. These models are computationally expensive, often taking up a large amount of memory and a long training period.  
As a result, the computational cost is not affordable to most researchers, and the long training time causes inconvenience for in-depth research on speech representation models in academia. A common practice is to download publicly accessed speech-based SSL models released by companies~\cite{hsu2021hubert, chen2022wavlm},  and then fine-tune them on the specific task. This poses challenges for researchers to improve and explore the pre-training of SSL models.
Therefore, it is essential to develop more efficient training for speech-based SSL models.

\begin{figure}[t] 
\centering
\center{\includegraphics[width=1.0\linewidth]{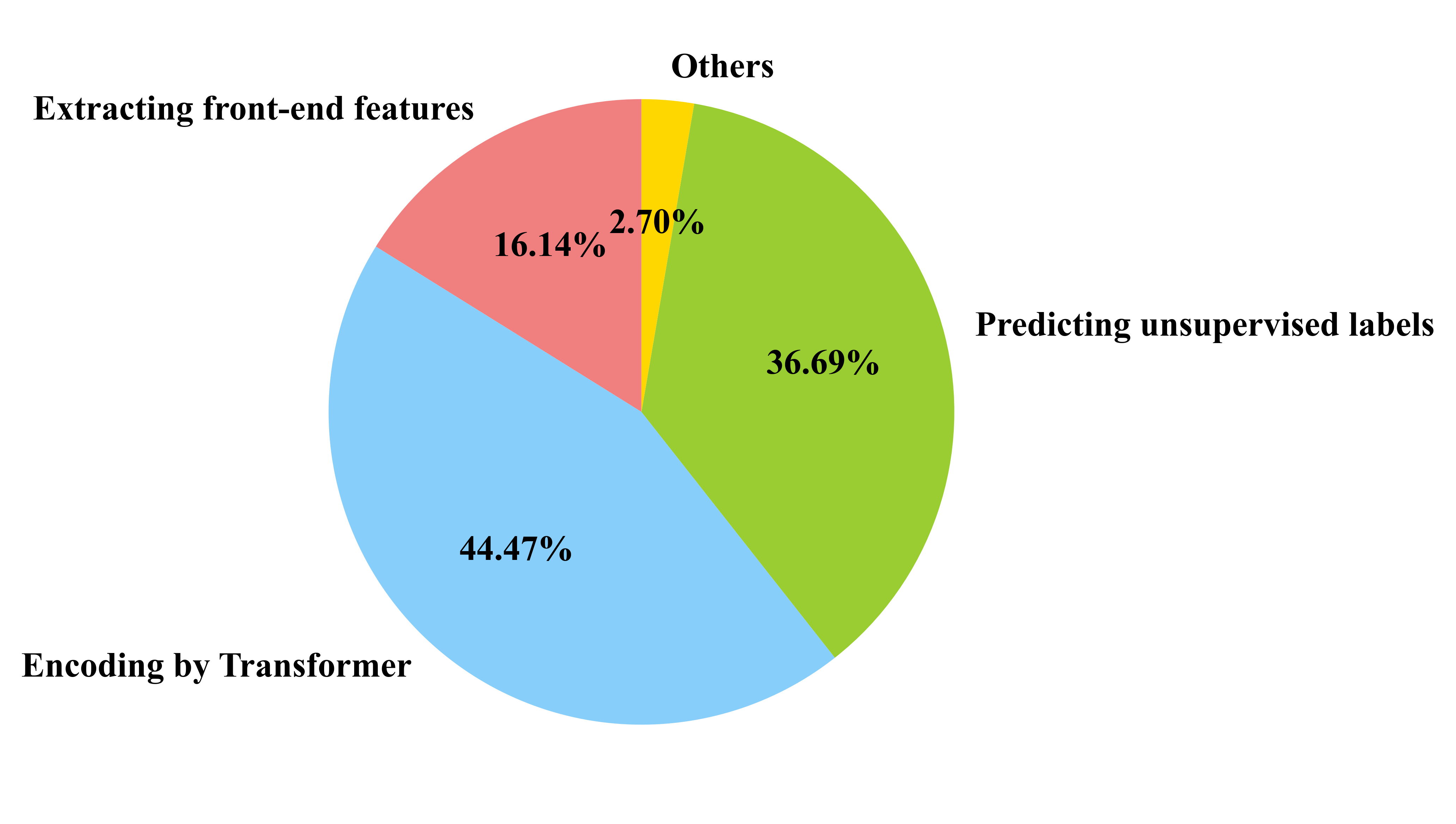}} 
\caption{The proportion of average time consumption at each part during the forward propagation process of the official HuBERT implementation in Fairseq.}
\label{fig:pie}
\end{figure}

In this paper, we choose the HuBERT model to analyze and optimize due to its promising performance and widespread application. 
Figure~\ref{fig:pie} displays the time consumption for each stage during the forward propagation process. 
The training process of the HuBERT model can be divided into four main parts, which are extracting front-end features from the raw audio waveforms, encoding the features by Transformer encoder, predicting the unsupervised labels, and other operations such as random masking. The distribution of time spent on each part provides a clear roadmap for us to further optimize the training efficiency. 
Specifically, we replace the raw waveform with the filter bank (Fbank) feature as front-end input to reduce the number of convolutional layers, which saves time for the online feature extraction. By further downsampling the Fbank features, the 
computational cost of the Transformer is significantly reduced. Besides, we simplify the original complicated HuBERT loss function by applying a standard cross-entropy loss, reducing the time for the final predicting stage. 
Our proposed optimized model, Fast-HuBERT, improves the training speed by 5.2x against HuBERT, while maintaining the performance of the model. 
Finally, we exploit the usage of unsupervised phoneme units in MonoBERT~\cite{ma2023pushing} and intermediate layer supervision (ILS) mechanism \cite{wang2021self}, which are able to boost the model performance without affecting training efficiency. 
\section{Background}
\subsection{Overview of HuBERT}
\label{sec:format}

\begin{figure*}[htb] 
\center{\includegraphics[width=1.0\textwidth]{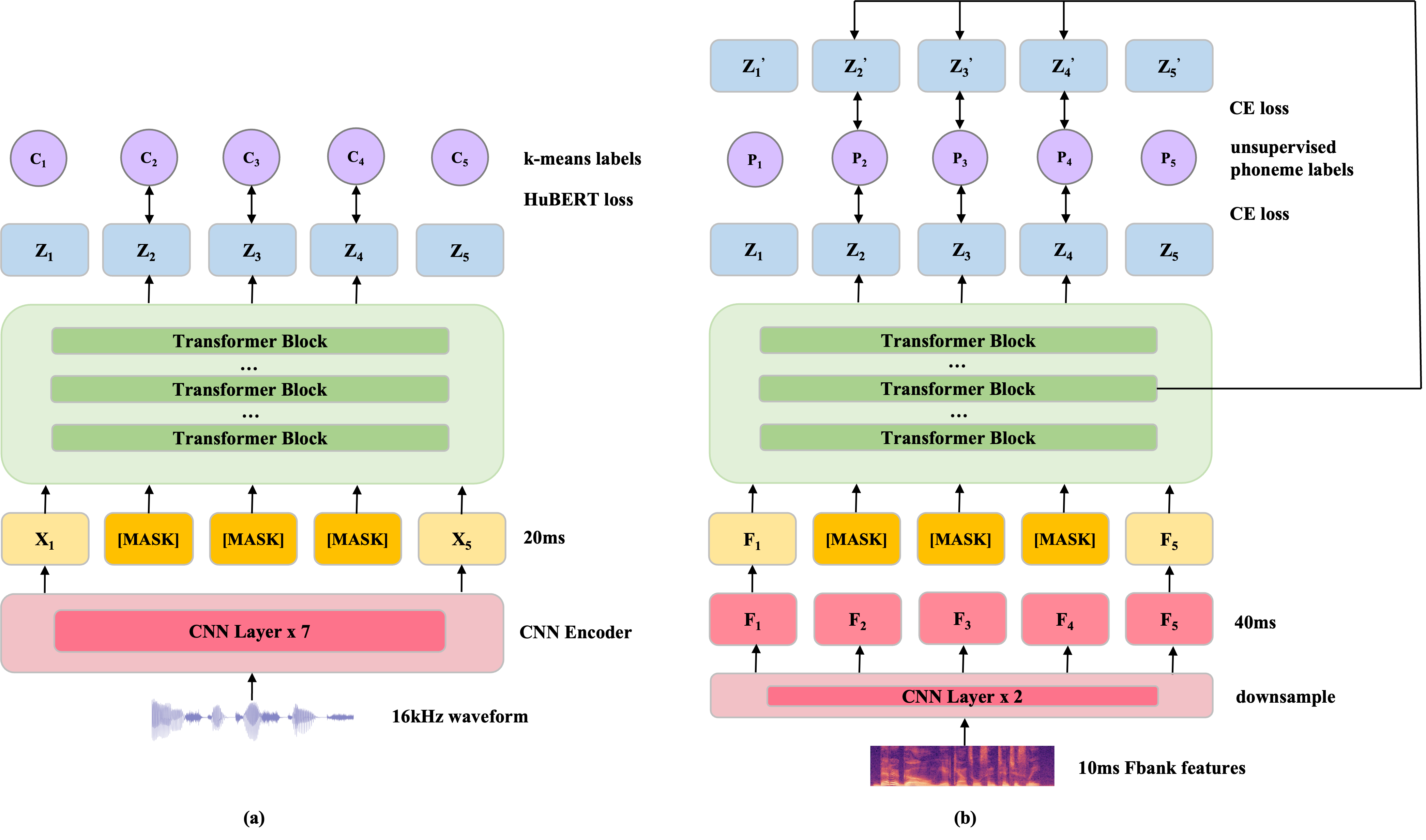}} 
\caption{(a) illustrates the original HuBERT model architecture. (b) presents the proposed Fast-HuBERT model architecture. Fast-HuBERT adopts downsampled Fbank features with a larger frameshift (e.g. 40ms) as inputs instead of features derived from raw waveforms. Besides, Fast-HuBERT applies a simplified cross-entropy loss instead of the original HuBERT loss as shown in Equation \ref{eq:2}, which significantly improves the pretraining efficiency. We also utilize unsupervised phoneme labels and the ILS mechanism to further improve model performance. }
\label{fig:model}
\end{figure*}

In this work, HuBERT \cite{hsu2021hubert} is selected as the representative speech-based SSL model for efficiency optimization. 
The overall model architecture is shown in Figure \ref{fig:model}(a). HuBERT, short for Hidden-Unit BERT, takes in raw speech waveforms and outputs predictions for k-means clustering labels corresponding to masked time steps. The waveform encoder consists of seven convolutional layers, which accept audios sampled at 16kHz and output feature sequences with a frameshift of 20ms. Next, HuBERT randomly selects 8\% of the time steps as the starting time step for a span of masks, and 10 consecutive time steps from that time step will be masked. 
Mask segments may have overlapping parts and the actual masked time steps are about 53\% of the total sequence. 

The speech features of the masked time steps are replaced with a learnable mask embedding. Finally, the predicted probability distribution over different clusters is computed by:
\begin{equation}
P_f(c|\widetilde{X},t) = \frac{\exp(sim(Ao_t,e_c)/\tau)}{ \sum^C_{c'=1} \exp(sim(Ao_t,e_{c'})/\tau) }
\label{eq:1}
\end{equation}
where $\widetilde{X}$ is the masked input, $[o_1,...,o_T]$ is the output sequence of the Transformer encoder, $A$ is the projection matrix, $e_c$ is the embedding for label c, $sim(\cdot,\cdot)$ computes the cosine similarity between two vectors, and $\tau$ scales the logit, which is set to 0.1. The k-means clusterings used in HuBERT can be derived from the 39-dimensional MFCC features and hidden features extracted from the Transformer intermediate layer of a pre-trained model.

\subsection{Related Work} 
Recent studies on self-supervised speech representation learning mainly focus on improving model performance, but few studies have explored methods for improving training efficiency, which is also an important issue. 

Many works have been proposed to optimize the HuBERT model from different perspectives. Unsupervised labels with higher quality can better guide the model’s pretraining. Some effective methods have been proposed to improve the label quality for HuBERT, such as HuBERT-AP~\cite{ren2022speech}, PBERT~\cite{wang2022supervision}, CTCBERT~\cite{fan2023ctcbert}.

MonoBERT~\cite{ma2023pushing} utilizes phoneme units generated by the generator of a pre-trained wav2vec-U 2.0~\cite{liu2023towards} model as labels. Wav2vec-U 2.0, optimized based on wav2vec-U~\cite{baevski2021unsupervised}, adopts an adversarial learning approach to train an end-to-end unsupervised phoneme recognition model, where the generator is trained to generate phoneme sequences that are as realistic as possible. The phoneme sequences are closer to the audio waveforms than the k-means clustering labels and contain more audio content information, which is beneficial for HuBERT pre-training. 
Wang et al.~\cite{wang2021self} proposed the Intermediate Layer Supervision mechanism for self-supervised learning (ILS-SSL).
In addition to a self-supervised loss on the top layer, ILS-SSL computes an auxiliary loss on the intermediate layers of Transformer, so that the lower layers are forced to learn more speech content information. This method is simple but effective, outperforming HuBERT significantly. 
 
There are also some recent works aiming to address the issue of high computational cost in model pre-training.
A related work MelHuBERT~\cite{lin2022melhubert} proposes to use Mel spectrograms as input and replace the loss function with the regular cross entropy to reduce computation and memory usage. Despite the similarities in terms of front-end optimization and loss function optimization, there are still several distinct differences. 
First, they only consider Fbank features of 10ms and 20ms frameshifts, which do not result in significant speed improvement, while we adopt a larger frameshift with a well-matched finetune strategy and noticeably improve the training efficiency. 
Besides, their implementation of feature downsampling and many other experimental settings are different from ours, for they focus on phoneme recognition and speaker identification tasks and evaluate on SUPERB~\cite{yang2021superb}, while we focus on ASR performance on the Librispeech~\cite{panayotov2015librispeech} benchmark.
Finally, they do not probe speed improvement results, while we not only present overall training acceleration results but also analyze the specific acceleration effects of each individual component.
Another related work data2vec 2.0~\cite{baevski2023efficient} improves the training efficiency of data2vec~\cite{baevski2022data2vec} by only encoding unmasked tokens, using a lightweight decoder and reusing target representations. It achieves the same performance as wav2vec 2.0 in 10.6x less time on Librispeech speech recognition task.

\section{Proposed Fast-HuBERT}
\label{sec:pagestyle}

In this section, we will elaborate on the proposed Fast-HuBERT, an efficient HuBERT training framework. The overall model architecture is presented in Figure \ref{fig:model}(b), in contrast to the original HuBERT in Figure \ref{fig:model}(a).

\subsection{Fbank front-end features}
\label{ssec:subhead}
\subsubsection{Analysis}
The original HuBERT model feeds raw audio sampled at 16 kHz into a convolutional waveform encoder to generate a feature sequence with a frameshift of 20 ms, which is aligned with target labels. According to the time analysis in Figure~\ref{fig:pie}, HuBERT spends a large proportion of time extracting features from the audio on-the-fly. 
Therefore, it is necessary to optimize the front-end feature extraction process. 

Here we remove the convolutional waveform encoder, extract Fbank features offline in advance, and use the features directly as model input. 
The frameshift of the Fbank features is set to 10ms, and it can be further downsampled in different scales, such as 20ms, 40ms, and 80ms. The downsampling module consists of a stack of one-dimensional convolutional layers followed by a non-linear activation via gated linear units (GLU). 

Due to the replacement of the front-end features, corresponding adjustments on the masking strategy are conducted, during both pre-training and fine-tuning phases. 
Since HuBERT uses raw waveforms as model input, multiple CNN layers can be viewed as a kind of learnable feature extractor. The masking strategy in HuBERT is conducted after CNN layers, which we call Post-Masking. 
In this work, since Fast-HuBERT uses Fbank features as input, we directly perform masking on the spectrogram before feeding it into CNN layers for downsampling, which we call Pre-Masking. In the pre-training phase, masking is utilized for the MLM task. While in the fine-tuning phase, masking is employed for data augmentation, just like that of SpecAugment~\cite{park2019specaugment}. 

\subsubsection{Pre-training with different frameshifts }
\label{sssec:subsubhead}

When using Fbank features as input and setting 20ms as frameshift for the Transformer encoder, one downsampling layer based on CNN is adopted. 
This operation causes a relative pre-training time reduction of 12.8\%, along with an additional 3.8\% relative WER improvement, compared to the original HuBERT. The replacement of front-end features is beneficial for both efficiency and performance.

The Transformer encoder~\cite{vaswani2017attention} has quadratic memory and computational complexity for its self-attention mechanism, depending on the length of the input sequence. 
We further change the frameshift from 20ms to 40ms, which is expected to significantly speed up the computation of the Transformer encoder, given the halved input feature sequence.  According to our experiments,
using 40ms frameshift features as Transformer input can result in a 3.3x speedup in the pre-training phase. Furthermore, the memory can hold more speech inputs because of the reduction in input sequence length. By simply enlarging the batch size, the pre-training speed increases by 5.2x.

More aggressively, we also attempt to use a frameshift of 80ms as input to the Transformer encoder. 
This achieves a much larger speedup in the pre-training stage, with 6.4x faster than the original HuBERT even without expanding the batch size. In this way, the model suffers from certain performance degradation on downstream ASR tasks. 
The result can be explained by the fact that a frameshift of 80ms is too large to accommodate the content information from the waveform. 
Using subword tokens during the fine-tuning phase can alleviate the problem caused by large frameshift, which is discussed in Section~\ref{subsubsec:subword}.

\subsubsection{Fine-tuning with subword transcriptions}
\label{subsubsec:subword}
Utilizing connectionist temporal classification (CTC)~\cite{graves2006connectionist} loss for downstream ASR finetuning task is common practice in the field of Speech SSL. In CTC, there is an assumption that the length of input speech features should be longer than that of target labels.
However, when we apply the input feature sequence with a frameshift of 80ms, the input sequence might be shorter than the target character sequence, which violates the key assumption lied in the CTC criterion. 
This mismatch prevents the model from being trained correctly, resulting in an abnormally high WER of over 80\%. 
In order to address this potential issue, we adopt the BPE (Byte Pair Encoder)~\cite{gage1994new} units as targets in the finetuning stage, instead of character units as in most previous work.

Inspired by this result, considering that the length of the 40ms frameshift feature sequence is shortened to one-half of its original size, we also try using the BPE subwords as labels. The experiment results show that using subwords instead of characters as labels distinctly brings an additional moderate WER reduction.

\begin{table*}[h!]

\centering
\caption{ASR results and pre-training speedup ratios of Fast-HuBERT on Librispeech dev/test sets. All models were pre-trained on LibriSpeech-960h, fine-tuned on Librispeech-100h, without a language model.}

\resizebox{\textwidth}{!}{
\begin{tabular}{c c c c c c c c c c }
\toprule
Model  & Input & Frameshift & Loss function & Finetune label & dev-clean &dev-other &test-clean &test-other &speedup \\
\midrule
HuBERT  & waveform & 20ms &HuBERT  loss & letter &5.4 & 13.1 & 5.6 & 12.8 & 1x \\
S1 & Fbank & 20ms & HuBERT loss & letter   &5.3 & 12.6 & 5.5 & 12.5 &1.1x \\
S2 & waveform  & 20ms &  CE loss & letter  &5.3 &11.9 &5.5 &11.9  &1.3x \\
S3 & Fbank & 20ms &  CE loss  & letter  &5.1 & 11.8 &5.3 &11.8 &1.6x \\
S4 & Fbank & 40ms &  CE loss  & letter  &6.1 & 13.8 & 6.3 & 13.9 &5.2x \\

S5 & Fbank & 40ms & CE loss & subword & 5.8 & 12.5 & 6.0 & 12.6 &5.2x \\
S6 & Fbank & 80ms & CE loss & subword & 7.5 & 15.3 & 7.3 & 15.6 &6.4x \\

\bottomrule
\end{tabular}}
\label{fig:asr_result}
\end{table*}

\subsection{Simplified cross entropy loss }
\label{sssec:subsubhead}
According to the elapsed time analysis shown in Figure \ref{fig:pie}, the computation of the loss function in HuBERT is another bottleneck due to its product operation between the output features and codebooks as shown in Equation \ref{eq:1}. 
In this work, we use the standard cross entropy loss instead, which treats HuBERT training as a common classification task. Specifically, an output feature sequence $[o_1,...,o_T]$ from the Transformer encoder is directly projected into logits $z$ by a linear fully connected layer, whose dimension is equal to the number of the k-means clusterings categories, denoted by a projection matrix $A$. Then, the logits are scaled by a temperature parameter $\tau$. Finally, the predicted probability distribution for each category is obtained by a softmax function, and the loss is calculated by a standard cross entropy function. The probability of each frame output $o_t$ being classified to the cluster $c$ is defined as:

\begin{align}
\label{eq:2}
& p_f(c|\mathbf{o_t}) = \frac{\exp(z_c)}{ \sum^C_{i=1} \exp(z_i)} \\ 
& \textbf{z} = A\mathbf{o_t}/\tau = (z_1,z_2,...,z_C) \in R^C
\end{align}

Such modification eliminates the need to update the embedding of each codeword and calculate cosine similarity. This simplification brings about an additional 23.2\% reduction in pre-training time compared to the original HuBERT, without hurting performance.

\begin{table*}[h!]
\centering
\caption{Ablation results of Fast-HuBERT with unsupervised phoneme labels and adding intermediate layer supervision.}

\begin{tabular}{l c c c c c }
\toprule
Model  &dev-clean &dev-other &test-clean &test-other&speedup   \\
\midrule
HuBERT  &5.4 & 13.1 & 5.6 & 12.8 & 1x \\
S5 & 5.8 & 12.5 & 6.0 & 12.6 &5.2x \\
 \quad + Unsupervised Phoneme Labels (S7) & 5.2 & 10.9 & 5.2 & \textbf{10.8}  &5.2x\\
    \qquad  \quad+ Intermediate Layer Supervision \textbf{(S8)} & \textbf{4.9} & \textbf{10.7} & \textbf{5.1} & 10.9 &\textbf{5.2x}\\
\bottomrule
\end{tabular}
\label{fig:ablation}

\end{table*}

\begin{table*}[h!]
\centering
\caption{Average time consumption (seconds/200 updates) of individual components in the forward propagation process.}

\begin{tabular}{c c c c c }
\toprule
Model  & Feature extraction & Transformer encoding  & Loss calculation & Others    \\
\midrule
HuBERT  &12.5 & 34.5 & 28.4 & 2.1 \\
S8 & 0.6 & 23.0 & 0.4 &2.5 \\
Speedup ratios & 95.2\%  & 33.3\% & 98.6\% & \\

\bottomrule
\end{tabular}
\label{fig:component}
\end{table*}

\section{Experiments}
\subsection{Experimental Setup}
\quad\ \ \textbf{Dataset.} We use the full 960h audios of Librispeech~\cite{panayotov2015librispeech} dataset for pre-training and train-clean-100 subset for fine-tuning. The standard dev (dev-clean and dev-other) and test (test-clean and test-other) data sets are adopted for performance evaluation.

\textbf{K-means clustering labels.} Following~\cite{hsu2021hubert}, we perform k-means clustering on 39-dimensional MFCC features to generate labels with 100 categories for the first pretraining iteration. Then, we run k-means clustering on the latent features extracted from the 6th transformer layer of the pretrained HuBERT model of the first iteration, to generate labels with 500 categories for our pre-training experiments.

\textbf{Unsupervised phoneme labels.} Following~\cite{ma2023pushing}, we utilize a pre-trained wav2vec-U 2.0 model and adjust the stride of its CNN layers to 1 to obtain frame-level aligned labels, which includes 40 categories in total.

\textbf{Pre-training.} During the pre-training stage, we adopt the same configuration for both HuBERT and Fast-HuBERT for a fair comparison. The models are pre-trained with 400k steps. The learning rate increases linearly from 0 to the peaking rate of 0.0005 in the first 32k steps and then decays linearly to zero during the remaining training time. We use Adam optimizer with $\beta=(0.9,0.98)$. 8\% frames of the extracted features are randomly selected as the start frame of a mask span, and 10 consecutive frames are masked. Our models are pre-trained on 8 32GB V100 GPUs, and at most 87.5 seconds of speech audio is contained in a batch on each GPU when the frameshift is set to 20ms.

\textbf{Fine-tuning with character labels.} In the finetuning stage, we finetune the pretrained model for 80k steps. The learning rate ramps from 0 to the peaking rate of 0.00003 in the first 8k steps, hold for 32k steps, and decays exponentially to 5\% of the peaking rate in the remaining 40k steps. The freeze-step hyperparameter is set to 10k steps, during which the Transformer parameters are frozen, and only the final projection matrix is updated. The WER result on the dev-other subset is used for model selection. In general, our model is fine-tuned on a single GPU and the update frequency is set to 8. For both raw waveforms and Fbank features, at most 200 seconds of speech audio is contained in a batch.

\textbf{Fine-tuning with subword labels.} When using Fbank features with large frameshifts as front-end input, we use subword labels in the fine-tuning stage. We use Hugging Face tokenizers package\footnote{https://huggingface.co/docs/tokenizers/index} to train a WordPiece tokenizer on Librispeech language model corpus. Then, we use the tokenizer to encode the transcriptions of validation and test datasets and use the subwords instead of letters as labels for fine-tuning. 

\textbf{Decoding.} We use the Viterbi algorithm with a beam size of 50 for decoding on the dev and test sets.

\subsection{Experimental Results}

Our code implementation is based on the fairseq~\cite{ott2019fairseq} framework.
Table~\ref{fig:asr_result} presents the WER results of the original HuBERT and its variants. The acceleration ratios of the pre-training speeds of different model structures compared with the original HuBERT model can be found in the last column. First, we use 20ms frameshift Fbank features and remove the convolutional waveform encoder (S1 model), which results in both pre-training speed acceleration and WER reductions on all test sets. Second, only replacing the HuBERT loss function with simplified cross-entropy loss (S2 model) can increase the pre-training speed by 1.3x and reduce the word error rates relatively by 1.9\%, 9.2\%, 1.8\%, and 7.0\%, respectively. Then, we combine both optimization methods (S3 model) and observe 1.6x pre-training acceleration and obvious performance improvements, which are relative reductions of 5.6\%, 9.9\%, 5.4\%, and 7.8\% in WER respectively. Next, we increase the frameshift to 40ms by downsampling Fbank features (S4 model) and achieve a remarkable 5.2x pre-training acceleration. However, due to the reduction of audio information, we observe certain degradation of model performance on four test sets. To address this issue, we change the labels in fine-tuning stage from letters to subwords (S5 model) and successfully make the WER lower than the original HuBERT model on both dev-other and test-other subsets, while the WER results on the two clean subsets remain slightly higher. Finally, we further increase the frameshift of the features to 80ms (S6 model) and achieve a significant 6.4x pre-training acceleration along with moderate WER degradation.  Therefore, a frameshift of 40ms is chosen in the proposed Fast-HuBERT.

\subsection{Ablation Study}

Leveraging the above optimization methods introduced for HuBERT,
we are able to pre-train a HuBERT model in about one day on 8 V100 GPUs, which allows various ideas to be explored in a controllable period. Based on this efficient training framework, we first explore two well-studied strategies to validate whether the proposed Fast-HuBERT accommodates well to these techniques, which are unsupervised phoneme target~\cite{ma2023pushing} and intermediate layer supervision~\cite{wang2021self}.

Table~\ref{fig:ablation} presents the WER and speed results of utilizing unsupervised phoneme labels and adding intermediate layer supervision (S8 model). These two techniques do not affect the pre-training time but can significantly improve the model performance. Specifically, we choose the 4th Transformer layer as the additional intermediate layer for loss calculation following ~\cite{wang2021self}. The WER is relatively reduced by 9.3\%, 18.3\%, 8.9\%, and 14.8\% respectively compared to the original HuBERT.

Furthermore, we conduct a quantitative comparison of the computation time between Fast-HuBERT and HuBERT for each training stage. 
Specifically, we calculate the average total time spent on each individual component in the forward propagation process every 200 training steps. The results are presented in Table~\ref{fig:component}. Experimental results show that after our optimization, the time spent on feature extraction and loss calculation steps is negligible, and the Transformer encoding time is reduced by 33.3\% due to the large frameshift.

\label{sec:typestyle}

\section{Conclusion}
In this work, we make an effort to optimize the training efficiency of self-supervised speech representation learning. Fast-HuBERT is developed to accelerate the training of the original HuBERT and maintain performance on downstream tasks such as speech recognition. The core optimization methods include Fbank features with larger frameshift, simplified cross-entropy loss, unsupervised phoneme labels, intermediate layer supervision mechanism in the pre-training stage, and subword labels in the fine-tuning stage. Experiments show that our proposed Fast-HuBERT can reduce the WER by 8.9\% and 14.8\% relatively on Librispeech test-clean and test-other subsets, along with 5.2x pre-training speedup.
In the future, we will apply these optimization methods to other speech-based SSL models to verify their versatilities and evaluate Fast-HuBERT on SUPERB to confirm its robustness. Besides, we will train LARGE Fast-HuBERT using the large-scale Libri-light dataset to further explore the capabilities of Fast-HuBERT.

\label{sec:majhead}

\section{Acknowledgements}
\vspace{-0.15cm}
This work was supported by the National Natural Science Foundation of China  (No. 62206171), the International Cooperation Project of PCL, and Alibaba Group through Alibaba Innovative Research Program.

\bibliographystyle{IEEEbib}
\bibliography{strings,refs}

\begin{thebibliography}{10}

\bibitem{e2ereview}
Jinyu Li,
\newblock ``Recent advances in end-to-end automatic speech recognition,''
\newblock {\em APSIPA Transactions on Signal and Information Processing}, 2022.

\bibitem{baevski2020wav2vec}
Alexei Baevski, Yuhao Zhou, Abdelrahman Mohamed, and Michael Auli,
\newblock ``Wav2vec 2.0: A framework for self-supervised learning of speech
  representations,''
\newblock {\em Proc. NeurIPS}, vol. 33, pp. 12449--12460, 2020.

\bibitem{hsu2021hubert}
Wei-Ning Hsu, Benjamin Bolte, Yao-Hung~Hubert Tsai, Kushal Lakhotia, Ruslan
  Salakhutdinov, and Abdelrahman Mohamed,
\newblock ``{HuBERT}: Self-supervised speech representation learning by masked
  prediction of hidden units,''
\newblock {\em IEEE/ACM TASLP}, vol. 29, pp. 3451--3460, 2021.

\bibitem{chen2022wavlm}
Sanyuan Chen, Chengyi Wang, Zhengyang Chen, Yu~Wu, Shujie Liu, Zhuo Chen, Jinyu
  Li, Naoyuki Kanda, Takuya Yoshioka, Xiong Xiao, et~al.,
\newblock ``Wavlm: Large-scale self-supervised pre-training for full stack
  speech processing,''
\newblock {\em IEEE Journal of Selected Topics in Signal Processing}, 2022.

\bibitem{ma2022mt4ssl}
Ziyang Ma, Zhisheng Zheng, Changli Tang, Yujin Wang, and Xie Chen,
\newblock ``{MT4SSL: Boosting Self-Supervised Speech Representation Learning by
  Integrating Multiple Targets},''
\newblock {\em Proc. InterSpeech}, 2023.

\bibitem{meng2022cobert}
Chutong Meng, Junyi Ao, Tom Ko, Mingxuan Wang, and Haizhou Li,
\newblock ``{CoBERT: Self-Supervised Speech Representation Learning Through
  Code Representation Learning},''
\newblock {\em Proc. InterSpeech}, 2022.

\bibitem{yang2021superb}
Shu-wen Yang, Po-Han Chi, Yung-Sung Chuang, Cheng-I~Jeff Lai, Kushal Lakhotia,
  Yist~Y Lin, Andy~T Liu, Jiatong Shi, Xuankai Chang, Guan-Ting Lin, et~al.,
\newblock ``{SUPERB: Speech processing Universal PERformance Benchmark},''
\newblock {\em Proc. InterSpeech}, 2021.

\bibitem{wang2023hubert}
Wei Wang and Yanmin Qian,
\newblock ``{HuBERT-AGG: Aggregated Representation Distillation of Hidden-Unit
  Bert for Robust Speech Recognition},''
\newblock in {\em Proc. ICASSP}, 2023.

\bibitem{chen2022large}
Zhengyang Chen, Sanyuan Chen, Yu~Wu, Yao Qian, Chengyi Wang, Shujie Liu, Yanmin
  Qian, and Michael Zeng,
\newblock ``Large-scale self-supervised speech representation learning for
  automatic speaker verification,''
\newblock in {\em Proc. ICASSP}, 2022.

\bibitem{ma2023leveraging}
Ziyang Ma, Wen Wu, Zhisheng Zheng, Yiwei Guo, Qian Chen, Shiliang Zhang, and
  Xie Chen,
\newblock ``{Leveraging Speech PTM, Text LLM, and Emotional TTS for Speech
  Emotion Recognition},''
\newblock {\em arXiv preprint arXiv:2309.10294}, 2023.

\bibitem{ma2023pushing}
Ziyang Ma, Zhisheng Zheng, Guanrou Yang, Yu~Wang, Chao Zhang, and Xie Chen,
\newblock ``Pushing the limits of unsupervised unit discovery for ssl speech
  representation,''
\newblock {\em Proc. Interspeech}, 2023.

\bibitem{wang2021self}
Chengyi Wang, Yu~Wu, Sanyuan Chen, Shujie Liu, Jinyu Li, Yao Qian, and Zhenglu
  Yang,
\newblock ``Self-supervised learning for speech recognition with intermediate
  layer supervision,''
\newblock in {\em Proc. ICASSP}, 2022.

\bibitem{ren2022speech}
Shuo Ren, Shujie Liu, Yu~Wu, Long Zhou, and Furu Wei,
\newblock ``Speech pre-training with acoustic piece,''
\newblock in {\em Proc. Interspeech}, 2022.

\bibitem{wang2022supervision}
Chengyi Wang, Yiming Wang, Yu~Wu, Sanyuan Chen, Jinyu Li, Shujie Liu, and Furu
  Wei,
\newblock ``Supervision-guided codebooks for masked prediction in speech
  pre-training,''
\newblock in {\em Proc. Interspeech}, 2022.

\bibitem{fan2023ctcbert}
Ruchao Fan, Yiming Wang, Yashesh Gaur, and Jinyu Li,
\newblock ``{CTCBERT}: Advancing hidden-unit {BERT} with {CTC} objectives,''
\newblock in {\em Proc. ICASSP}. IEEE, 2023.

\bibitem{liu2023towards}
Alexander~H Liu, Wei-Ning Hsu, Michael Auli, and Alexei Baevski,
\newblock ``Towards end-to-end unsupervised speech recognition,''
\newblock in {\em Proc. SLT}. IEEE, 2023.

\bibitem{baevski2021unsupervised}
Alexei Baevski, Wei-Ning Hsu, Alexis Conneau, and Michael Auli,
\newblock ``Unsupervised speech recognition,''
\newblock {\em Proc. NeurIPS}, vol. 34, pp. 27826--27839, 2021.

\bibitem{lin2022melhubert}
Tzu-Quan Lin, Hung-yi Lee, and Hao Tang,
\newblock ``{MelHuBERT}: A simplified {HuBERT} on mel spectrogram,''
\newblock in {\em Proc. ICASSP}, 2023.

\bibitem{panayotov2015librispeech}
Vassil Panayotov, Guoguo Chen, Daniel Povey, and Sanjeev Khudanpur,
\newblock ``Librispeech: An {ASR} corpus based on public domain audio books,''
\newblock in {\em Proc. ICASSP}. IEEE, 2015, pp. 5206--5210.

\bibitem{baevski2023efficient}
Alexei Baevski, Arun Babu, Wei-Ning Hsu, and Michael Auli,
\newblock ``Efficient self-supervised learning with contextualized target
  representations for vision, speech and language,''
\newblock in {\em Proc. ICML}. PMLR, 2023, pp. 1416--1429.

\bibitem{baevski2022data2vec}
Alexei Baevski, Wei-Ning Hsu, Qiantong Xu, Arun Babu, Jiatao Gu, and Michael
  Auli,
\newblock ``Data2vec: A general framework for self-supervised learning in
  speech, vision and language,''
\newblock in {\em Proc. ICML}, 2022, pp. 1298--1312.

\bibitem{park2019specaugment}
Daniel~S Park, William Chan, Yu~Zhang, Chung-Cheng Chiu, Barret Zoph, Ekin~D
  Cubuk, and Quoc~V Le,
\newblock ``{SpecAugment}: A simple data augmentation method for automatic
  speech recognition,''
\newblock in {\em Proc. Interspeech}, 2019.

\bibitem{vaswani2017attention}
Ashish Vaswani, Noam Shazeer, Niki Parmar, Jakob Uszkoreit, Llion Jones,
  Aidan~N Gomez, {\L}ukasz Kaiser, and Illia Polosukhin,
\newblock ``Attention is all you need,''
\newblock {\em Proc. NeurIPS}, vol. 30, 2017.

\bibitem{graves2006connectionist}
Alex Graves, Santiago Fern{\'a}ndez, Faustino Gomez, and J{\"u}rgen
  Schmidhuber,
\newblock ``Connectionist temporal classification: labelling unsegmented
  sequence data with recurrent neural networks,''
\newblock in {\em Proc. ICML}, 2006, pp. 369--376.

\bibitem{gage1994new}
Philip Gage,
\newblock ``A new algorithm for data compression,''
\newblock {\em C Users Journal}, vol. 12, no. 2, pp. 23--38, 1994.

\bibitem{ott2019fairseq}
Myle Ott, Sergey Edunov, Alexei Baevski, Angela Fan, Sam Gross, Nathan Ng,
  David Grangier, and Michael Auli,
\newblock ``fairseq: A fast, extensible toolkit for sequence modeling,''
\newblock {\em arXiv preprint arXiv:1904.01038}, 2019.

\end{thebibliography}

\end{document}